\documentclass[letterpaper]{article} 
\usepackage{aaai23}  
\usepackage{times}  
\usepackage{helvet}  
\usepackage{courier}  
\usepackage[hyphens]{url}  
\usepackage{graphicx} 
\usepackage{natbib}  
\usepackage{caption} 
\usepackage{amsmath}
\usepackage{amsmath}
\usepackage{graphicx}
\usepackage{amsmath}
\usepackage{amssymb}
\usepackage{booktabs}
\usepackage{multirow}
\usepackage{algorithm}
\usepackage{algorithmic}
\usepackage{xcolor}
\usepackage{graphicx}
\usepackage{algorithm}
\usepackage{algorithmic}
\usepackage{tikz}
\usepackage{comment}
\usepackage{amsmath,amssymb}
\usepackage{multirow}
\usepackage{newfloat}
\usepackage{listings}

\DeclareCaptionStyle{ruled}{labelfont=normalfont,labelsep=colon,strut=off} 
\floatstyle{ruled}
\newfloat{listing}{tb}{lst}{}
\floatname{listing}{Listing}
\title{Structure Flow-Guided Network for Real Depth Super-Resolution}
\author{
    Jiayi Yuan\equalcontrib, Haobo Jiang\equalcontrib, Xiang Li, Jianjun Qian, Jun Li\footnotemark[2], Jian Yang\thanks{corresponding authors}
}
\affiliations{
    PCA Lab, Key Lab of Intelligent Perception and Systems for High-Dimensional Information of Ministry of Education\\
    Jiangsu Key Lab of Image and Video Understanding for Social Security\\
    School of Computer Science and Engineering, Nanjing University of Science and Technology, Nanjing, China\\
     
\{jiayiyuan, jiang.hao.bo, xiang.li.implus, csjqian, junli, csjyang\}@njust.edu.cn
}

\usepackage{bibentry}

\begin{document}
\makeatletter
\let\@oldmaketitle\@maketitle
\renewcommand{\@maketitle}{\@oldmaketitle
\begin{center}
\includegraphics[width=0.92\textwidth]{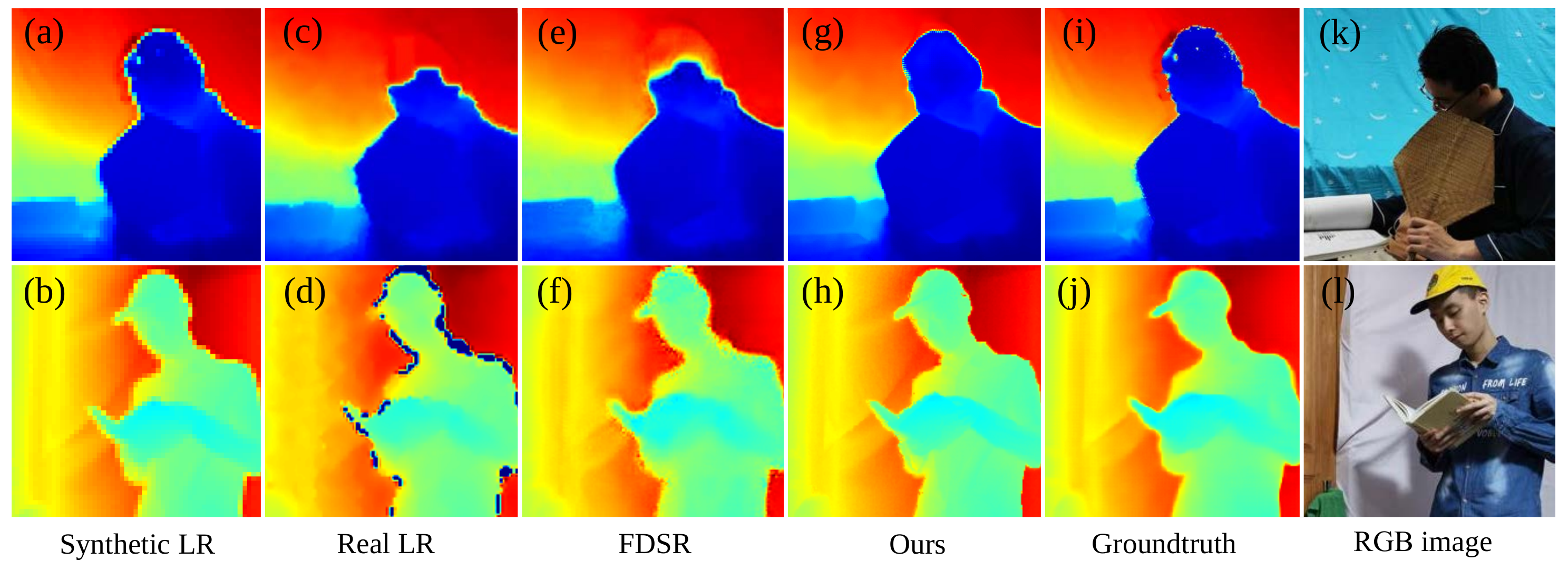}
\captionof{figure}{In this paper, we propose a novel structure flow-guided method for real-world DSR. Our method obtains better depth edge recovery ($g$-$h$), compared to ($e$) and ($f$) using the SOTA method, FDSR \cite{he2021towards}. ($a$-$b$) Synthetic LR depth maps; ($c$) Real LR depth map with the structural distortion; ($d$) Real LR depth map with the edge noise (e.g., holes); ($i$-$j$) Ground-truth HR depth maps; ($k$-$l$) RGB image guidance.}
 \label{fig:Fig1}
\end{center}
\bigskip}

\maketitle

\begin{abstract} 
Real depth super-resolution (DSR), unlike synthetic settings, is a challenging task due to the structural distortion and the edge noise caused by the natural degradation in real-world low-resolution (LR) depth maps. These defeats result in significant structure inconsistency between the depth map and the RGB guidance, which potentially confuses the RGB-structure guidance and thereby degrades the DSR quality. In this paper, we propose a novel structure flow-guided DSR framework, where a cross-modality flow map is learned to guide the RGB-structure information transferring for precise depth upsampling. Specifically, our framework consists of a cross-modality flow-guided upsampling network (CFUNet) and a flow-enhanced pyramid edge attention network (PEANet). CFUNet contains a trilateral self-attention module combining both the geometric and semantic correlations for reliable cross-modality flow learning. Then, the learned flow maps are combined with the grid-sampling mechanism for coarse high-resolution (HR) depth prediction. PEANet targets at integrating the learned flow map as the edge attention into a pyramid network to hierarchically learn the edge-focused guidance feature for depth edge refinement. Extensive experiments on real and synthetic DSR datasets verify that our approach achieves excellent performance compared to state-of-the-art methods. 
\end{abstract}

\section{Introduction} 
With the fast development of cheap RGB-D sensors, depth maps have played a much more important role in a variety of computer vision applications, such as object recognition \cite{blum2012learned,eitel2015multimodal}, 3D reconstruction \cite{hou20193d,newcombe2011kinectfusion}, and virtual reality \cite{meuleman2020single}). 
However, the defects (e.g., low resolution and structural distortion) lying in the cheap RGB-D sensors (e.g., Microsoft Kinect and HuaweiP30Pro), still hinder their more extensive applications in real world. 
Also, although the popular DSR methods \cite{song2020channel,kim2021deformable,sun2021learning} have achieved excellent DSR accuracy on synthetic LR depth maps, the significant domain gap between the real and the synthetic data largely degrades their DSR precision on the real data. 

This domain gap is mainly caused by different generation mechanisms of the LR depth map. 
The synthetic LR depth map is usually generated via artificial degradation (e.g., down-sampling operation), while the real one is from natural degradation (e.g., noise, blur, and distortion). 
Different from the synthetic data, there are two challenges of the real-data DSR as below. 
The first one is the severe structural distortion (see Fig.~\ref{fig:Fig1} (c)), especially for the low-reflection glass surface or the infrared-absorbing surface. 
The second one is the edge noise even the holes (see Fig.~\ref{fig:Fig1} (d)), caused by the physical limitations or the low processing power of the depth sensors. 
Both of the challenges above present a significant difference between the real and the synthetic data, which inherently degrades the generalization precision of the synthetic DSR methods to the real data. 

In this paper, we develop a novel structure flow-guided DSR framework to handle the above challenges. 
For the structural distortion, we propose a cross-modality flow-guided upsampling network (CFUNet) that learns a structured flow between the depth map and the RGB image to guide their structure alignment for the recovery of the distorted depth structure. 
It includes two key components: a trilateral self-attention module and a cross-modality cross-attention module. 
In detail, the former leverages the geometric and semantic correlations (i.e., coordinate distance, pixel difference, feature difference) to guide the relevant 
depth-feature aggregation into each depth feature to supplement the missing depth-structure information. 
The latter utilizes the enhanced depth feature and the RGB feature as the input for their sufficient message passing and flow-map generation. 
Finally, we combine the flow map with the grid-sampling mechanism for the coarse HR depth prediction. 

For the edge noise, we present a flow-enhanced pyramid edge attention network (PEANet) that integrates the learned structure flow map as the edge attention into a pyramid network to learn the edge-focused guidance feature for the edge refinement of the coarse HR depth map predicted above. 
Considering the structure clue (i.e., edge region tends to own significant flow-value fluctuations) lying in the learned flow map, we combine the flow map with the RGB feature to form the flow-enhanced RGB feature for highlighting the RGB-structure region. 
Then, we feed the flow-enhanced RGB feature into an iterative pyramid network for its edge-focused guidance feature learning. The low-level guidance features effectively filter the RGB-texture noise (guided by the flow map), while the high-level guidance features exploit the rich context information for more precise edge-feature capture. 
Finally, we pass the learned guidance feature and the depth feature into a decoder network to predict the edge-refined HR depth map. 
Extensive experiments on challenging real-world datasets verify the effectiveness of our proposed method (see examples in Fig.~\ref{fig:Fig1}(g-h)). In summary, our contributions are as follows: 
\begin{itemize}
\item We propose an effective cross-modality flow-guided upsampling network (CFUNet), where a structure flow map is learned to guide the structure alignment between the depth map and the RGB image for the recovery of the distorted depth edge. 
\item We present a flow-enhanced pyramid edge attention network (PEANet) that integrates the flow map as edge attention into a pyramid network to hierarchically learn the edge-focused guidance feature for edge refinement. 
\item Extensive experiments on the real and synthetic datasets verify the effectiveness of the proposed framework, and we achieve state-of-the-art restoration performance on multiple DSR dataset benchmarks.
\end{itemize}

\begin{figure*}[t]
  \centering
     \includegraphics[width=0.98\linewidth]{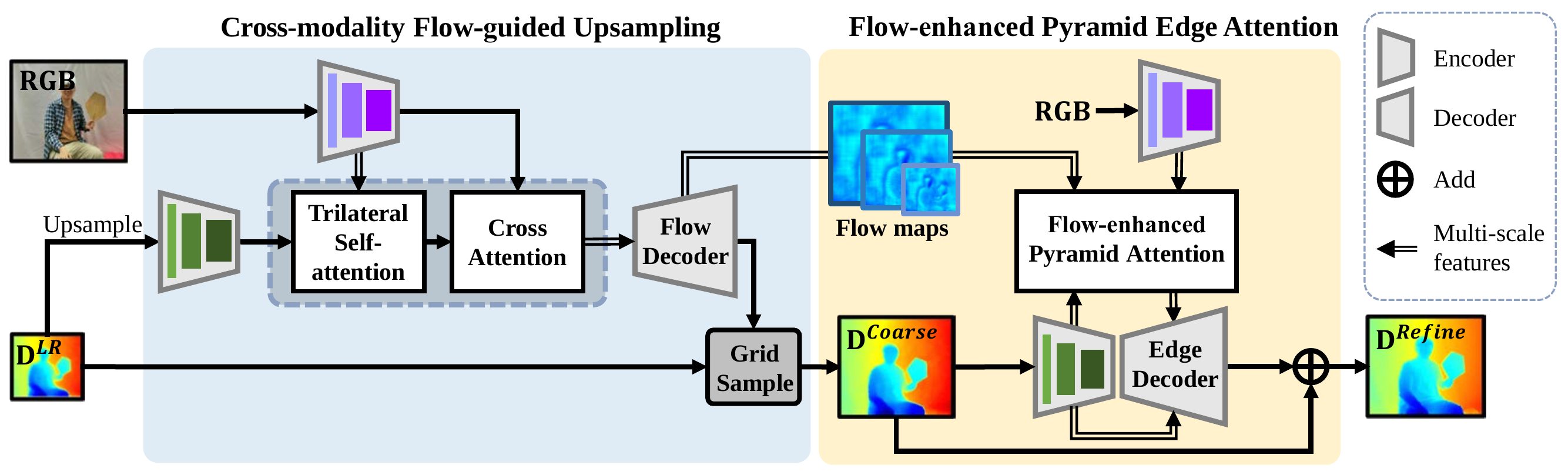}
   \caption{The pipeline of our structure flow-guided DSR framework. 
   Given the LR depth map and the RGB image, the left block (blue) first generates the flow maps through a trilateral self-attention module and a cross-attention module, and predicts the coarse depth map $\mathbf{D}^{coarse}$ with the flow-based grid-sampling. 
   Then, the right block (yellow) integrates the RGB/depth features and the flow map (as edge attention) to learn the edge-focused guidance feature for edge refinement ($\mathbf{D}^{refine}$).
   }
   \label{fig:framework}
\end{figure*}

\section{Related Work}
\subsection{Synthetic Depth Super-Resolution } 
The synthetic depth super-resolution (DSR) architectures can be divided into the pre-upsampling methods and the progressive upsampling methods~\cite{wang2020depth}. 
The pre-upsampling DSR methods first upsample the input depth with interpolation algorithms (e.g., bicubic) from LR to HR, and then feed it into depth recovery network layers. \cite{li2016deep} introduce the first pre-upsampling network architecture. As this method handles arbitrary scaling factor depth, more and more similar approaches have been presented to further facilitate DSR  task~\cite{li2019joint,lutio2019guided,zhu2018co,chen2018single,hao2019multi,su2019pixel}. 
However, upsampling in one step is not suitable for large scaling factors simply because it usually leads to losing much detailed information. 
To tackle these issues, a {progressive upsampling structure} is designed in MSG-net\cite{2016Multi}, which gradually upsamples the LR depth map by transposed convolution layers at different scale levels. Since then, various progressive upsample-based methods have been proposed that greatly promote the development of this domain\cite{2016Multi,2019Hierarchical,he2021towards,zuo2019multi}. 
Recently, the joint-task learning framework achieves impressive performance, such as DSR $\&$ completion \cite{yan2022learning}, depth estimation $\&$ enhancement \cite{Wang_2021_ICCV} and DSR $\&$ depth estimation \cite{tang2021bridgenet,sun2021learning}. 
Inspired by these joint-task methods, we combine the alignment task with the super-resolution task to distill the cross-modality knowledge for robust depth upsampling.
\subsection{Real-world Depth Super-Resolution}
In recent years, the super-resolution for real-world images has been under the spotlight, which involves upsampling, denoising, and hole-filling. Early traditional depth enhancement methods \cite{yang2014color, liu2016robust,liu2018depth} are based on complex and time-consuming optimization. For fast CNN-based DSR, AIR \cite{song2020channel} simulates the real LR depth map by combining the interval degradation and the bicubic degradation, and proposes a channel attention based network for real DSR. 
PAC \cite{su2019pixel} and DKN \cite{kim2021deformable} utilize the adaptive kernels calculated by the neighborhood pixels in RGB image for robust DSR. 
FDSR\cite{he2021towards} proposes the octave convolution for frequency domain separation, which achieves outstanding performance in real datasets.
Although these methods handle the large modality gap between the guidance image and depth map, the structure misalignment between the depth map and the RGB image still leads them to suffer from serious errors around the edge regions. 
Different from the general paradigms, we introduce a novel structure flow-guided framework, which exploits the cross-modality flow map to guide the RGB-structure information transferring for real DSR.

\section{Approach} 
In the following, we introduce our structure flow-guided DSR framework for robust real-world DSR. As shown in Fig.~\ref{fig:framework}, our framework consists of two modules: a cross-modality flow-guided upsampling network (CFUNet) and a flow-enhanced pyramid edge attention network (PEANet). 
Given an LR depth map $\mathbf{D}^{{LR}}\in \mathbb{R}^{H_0\times W_0}$ and its corresponding HR RGB image $\mathbf{I}\in \mathbb{R}^{H\times W \times 3}$ ($H/H_0=W/W_0=s$ and $s$ is the scale factor), CFUNet first learns the cross-modality flow to guide the structure alignment between depth the RGB for coarse HR depth prediction. 
Then, PEANet exploits the structure flow as edge attention to learn the edge-focused guidance feature for edge refinement.

\begin{figure*}[t]
  \centering
     \includegraphics[width=0.9\linewidth]{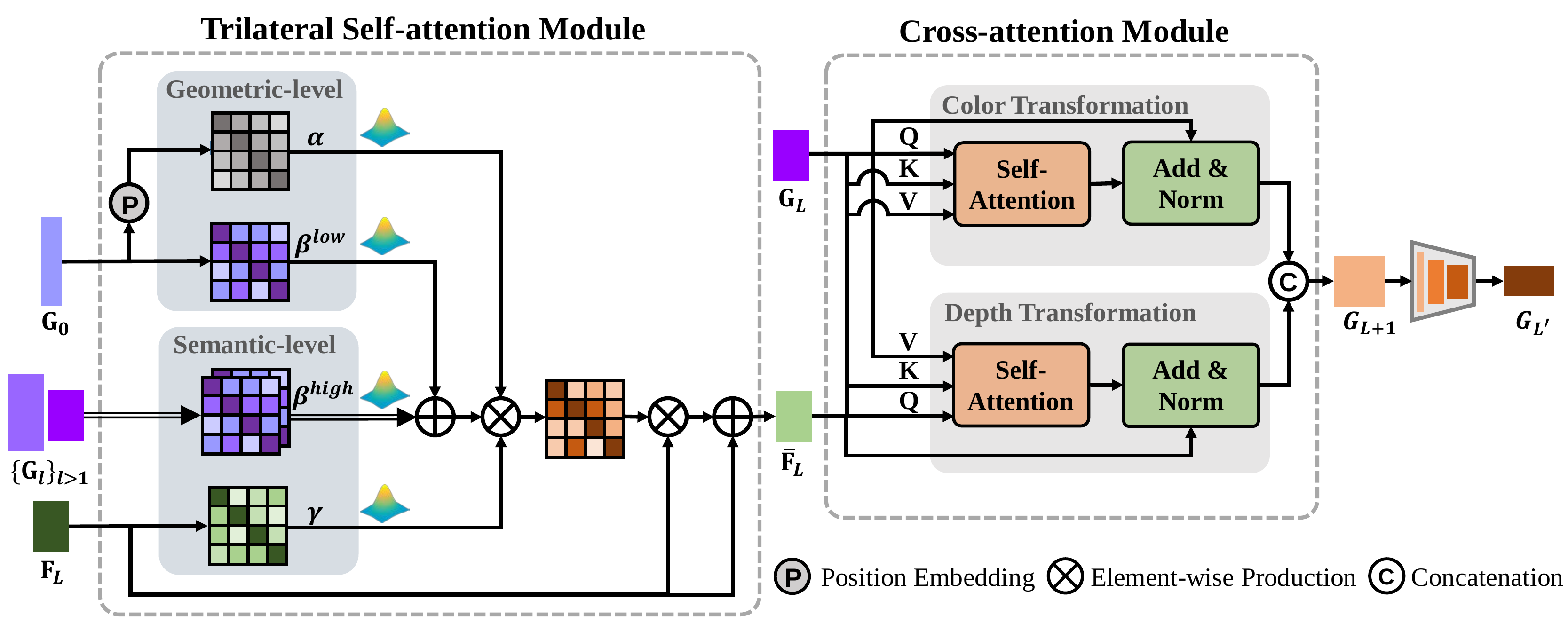}
   \caption{The architecture of the trilateral self-attention module and the cross-attention module.}
   \label{fig:upsample}
\end{figure*}
\subsection{Cross-modality Flow-guided Upsampling Network}
\label{sec:TF} 

As demonstrated in Fig.~\ref{fig:Fig1} (c), the structural distortion of the real LR depth map leads to the significant structure misalignment between the RGB image and the depth map, which potentially damages the structure guidance of RGB images for depth edge recovery. 
To handle it, our solution is to learn an effective cross-modality flow map between the depth and the RGB to identify their structure relationship. 
Then, guided by the learned flow map, we align the structure of the depth map to the RGB image for the recovery of the distorted depth edge. 
Next, we will describe our network in terms of the feature extraction, the trilateral attention-based flow generation, and the flow-guided depth upsampling. 

\textbf{Feature extraction.} 
To achieve the consistent input size, we first upsample the LR depth map $\mathbf{D}^{{LR}}$ to a resolution map $\mathbf{D}^{{Bic}}\in \mathbb{R}^{H\times W}$ with the bicubic interpolation. 

Then, we feed the upsampled depth map and the RGB image into an encoder for their feature extraction: $\{\mathbf{F}_l\in \mathbb{R}^{H\times W \times D}\}_{l=1}^L$ and $\{\mathbf{G}_l \in \mathbb{R}^{H\times W \times D}\}_{l=1}^L$ where the subscript $l$ denotes the feature output in $l$-th layer of the encoder. 

\textbf{Trilateral attention-based flow generation.} 
The key to generating a reliable cross-modality flow map is to model a robust relationship between the RGB and the depth map. 
Nevertheless, the serious structural distortion caused by the natural degradation potentially increases the modality gap between the depth and the RGB. Thereby, it's difficult to directly exploit a general attention mechanism to model such a relationship.
To mitigate it, we target at enhancing the depth feature through a proposed trilateral self-attention block so that the distorted depth-structure information can be largely complemented for relationship modeling. 
As shown in Fig.~\ref{fig:upsample}, our trilateral self-attention block fuses the geometric-level correlation and the semantic-level correlation to jointly guide the depth feature enhancement. 
It's noted that we just enhance the depth feature $\mathbf{F}_L$ in the last layer ($L$-th layer):
\begin{equation}\label{kernel}\small
\begin{split}
	\setlength{\abovedisplayskip}{1.2pt}
	\setlength{\belowdisplayskip}{1.2pt}
\bar{\mathbf{F}}_{L}^{(i)} = \sum_{j} \boldsymbol{\alpha}_{i,j}\cdot(\boldsymbol{\beta}^{low}_{i,j}+\boldsymbol{\beta}^{high}_{i,j})\cdot\boldsymbol{\gamma}_{i,j} \cdot \mathbf{F}_{L}^{(j)} + \mathbf{F}_{L}^{(j)},
\end{split}
\end{equation}
where $\mathbf{F}_{L}^{(j)}$ ($1\leq j \leq H\times W$) denotes the $j$-th depth-pixel feature and $\bar{\mathbf{F}}_{L}^{(i)}$ denotes the $i$-th enhanced depth feature ($1\leq i \leq H\times W$).
The geometric-level correlation contains a spatial kernel $\boldsymbol{\alpha} \in \mathbb{R}^{(H\times W)\times (H \times W)}$ and a low-level color kernel $\boldsymbol{\beta}^{low} \in \mathbb{R}^{(H\times W)\times (H \times W)}$, while the semantic-level correlation contains a high-level color semantic kernel $\boldsymbol{\beta}^{high} \in \mathbb{R}^{(H\times W)\times (H \times W)}$ and a depth semantic kernel $\boldsymbol{\gamma} \in \mathbb{R}^{(H\times W)\times (H \times W)}$.
In detail, we formulate the spatial kernel as a coordinate distance-aware Gaussian kernel:
\begin{equation}\small
\begin{split}
\boldsymbol{\alpha}_{i,j} &= \operatorname{Gaussian}(\|\operatorname{Coor}(i)-\operatorname{Coor}(j)\|_2, \sigma_s), 
\end{split}
\end{equation}
where $\operatorname{Gaussian}(x, \sigma)=\frac{1}{\sigma\sqrt{2\pi}}\exp{(-\frac{x^2}{2\sigma^2})}$ is the Gaussian function. $\operatorname{Coor}(i) \in \mathbb{R}^2$ denotes the row-column coordinates of pixel $i$ at the depth map and $\sigma_s$ is the kernel variance. The low-level and high-level color kernels are defined by the Gaussian kernels with the low-level and the semantic-level RGB feature similarity, whose kernel sum is:
\begin{align}\small
\boldsymbol{\beta}^{low}_{i,j}+\boldsymbol{\beta}^{high}_{i,j} &=
 {\sum_{l=0}^{L}}\operatorname{Gaussian}(\|\textbf{G}_l^{(i)}-\textbf{G}_l^{(j)}\|_2, \sigma_c).
\end{align}
The depth semantic kernel is designed based on the depth feature similarity in the $L$-th layer:
\begin{equation}\small
\begin{split}
\boldsymbol{\gamma}_{i,j} &=
\operatorname{Gaussian}(\|\textbf{F}_{L}^{(i)}-\textbf{F}_{L}^{(j)}\|_2,
\sigma_d).
\end{split}
\end{equation}
Guided by the geometric and semantic kernels above, the correlated depth information can be effectively aggregated into each depth feature through Eq.\ref{kernel} for depth feature completion and enhancement. 

Then, we feed the enhanced depth feature $\bar{\mathbf{F}}_{L}$ and the RGB feature ${\mathbf{G}}_{L}$ into the cross-attention block for their efficient cross-modality feature intersection:
\begin{equation}\small
	\setlength{\belowdisplayskip}{1pt}
	\begin{split}
		\Tilde{\mathbf{F}}_{L}^{(i)} &= \bar{\mathbf{F}}_{L}^{(i)} + \operatorname{MLP}(\sum_{j}\operatorname{softmax}_j(\phi_q(\bar{\mathbf{F}}_{L}^{(i)})^\top\phi_k({\mathbf{G}}_{L}^{(j)}))\phi_v({\mathbf{G}}_{L}^{(j)})), \\
		{\Tilde{\mathbf{G}}_{L}^{(i)}} &= {\mathbf{G}}_{L}^{(i)} + \operatorname{MLP}(\sum_{j}\operatorname{softmax}_j(\phi_q({\mathbf{G}}_{L}^{(i)})^\top\phi_k(\bar{\mathbf{F}}_{L}^{(j)})\phi_v(\bar{\mathbf{F}}_{L}^{(j)})),\\
	\end{split}
\end{equation}
where $\phi_q$, $\phi_k$ and $\phi_v$ are the projection functions of the query, the key and the value in our nonlocal-style cross-attention module. 
With the query-key similarity, the value can be retrieved for feature enhancement.
Then, we concatenate the enhanced depth feature $\Tilde{\mathbf{F}}_{L}$ and RGB feature $\Tilde{\mathbf{G}}_{L}$ and pass them into a multi-layer convolutional network to obtain their correlated feature at each layer $\{{\mathbf{G}_l} \}_{l=L+1}^{L'}$. 
Finally, following \cite{Dosovitskiy_2015_ICCV}, based on the previously extracted features $\{{\mathbf{G}_l} \}_{l=1}^{L}$ and the correlated features $\{{\mathbf{G}_l} \}_{l=L+1}^{L'}$, we  exploit a decoder network to generate the multi-layer flow maps $\{\boldsymbol{\Delta}_l\}_{l=1}^{L'}$, where the flow generation in layer $l$ can be formulated as:
\begin{equation} \label{flow0}
\begin{split}
\mathbf{G}_{l+1}^{{flow}}, \boldsymbol{\Delta}_{l+1} = \operatorname{deconv}(\operatorname{Cat}[\mathbf{G}_l^{{flow}}, \boldsymbol{\Delta}_{l}, \mathbf{G}_{L'-l-1}]),
\end{split}
\end{equation}
where $\mathbf{G}_l^{{flow}}$ denotes the intermediate flow feature and $\operatorname{deconv}$ consisting of a deconvolution operation and a convolutional block ($\mathbf{G}_{1}^{{flow}}, \boldsymbol{\Delta}_{1} = \operatorname{deconv}(\mathbf{G}_{L'})$).

\textbf{Flow-guided depth upsampling module.} 
With the learned flow map $\boldsymbol{\Delta}_{L'}$ in the last layer, we combine it with the grid-sampling strategy for the HR depth map prediction. 
In detail, the value of the HR depth map is the bilinear interpolation of the neighborhood pixels in LR depth map $\mathbf{D}^{{LR}}$, where the neighborhoods are defined according to the learned flow field, which can be formulated as:
\begin{equation}
\begin{split}
\mathbf{D}^{{coarse}}=\operatorname{Grid-Sample}(\mathbf{D}^{{LR}},\boldsymbol{\Delta}_{L'}),
\end{split}
\end{equation}
where $\operatorname{Grid-Sample}$ denotes the upsampling operation computing the output using pixel values from neighborhood pixels and pixel locations from the grid \cite{li2020semantic}.

\subsection{Flow-enhanced Pyramid Edge Attention Network}
\label{sec:UR}
In order to further improve our DSR precision in the case of the edge noise problem, we propose a flow-enhanced pyramid network, where the learned structure flow is served as the edge attention to hierarchically mine edge-focused guidance feature from the RGB image for the edge refinement of $\mathbf{D}^{{coarse}}$. 
Specifically, we first feed the previously predicted HR depth map $\mathbf{D}^{coarse}$ and the RGB image into an encoder network to extract their features: $\{{\mathbf{F}}^{coarse}_t\}_{t=1}^{T+1}$ and $\{{\mathbf{G}}_t\}_{t=1}^{T}$, where subscript $t$ indicates the extracted feature at the $t$-th layer. 
Then, we propose the flow-enhanced pyramid attention module and the edge decoder module as follows for refined HR depth prediction.

\begin{figure}[t]
  \centering
     \includegraphics[width=0.95\linewidth]{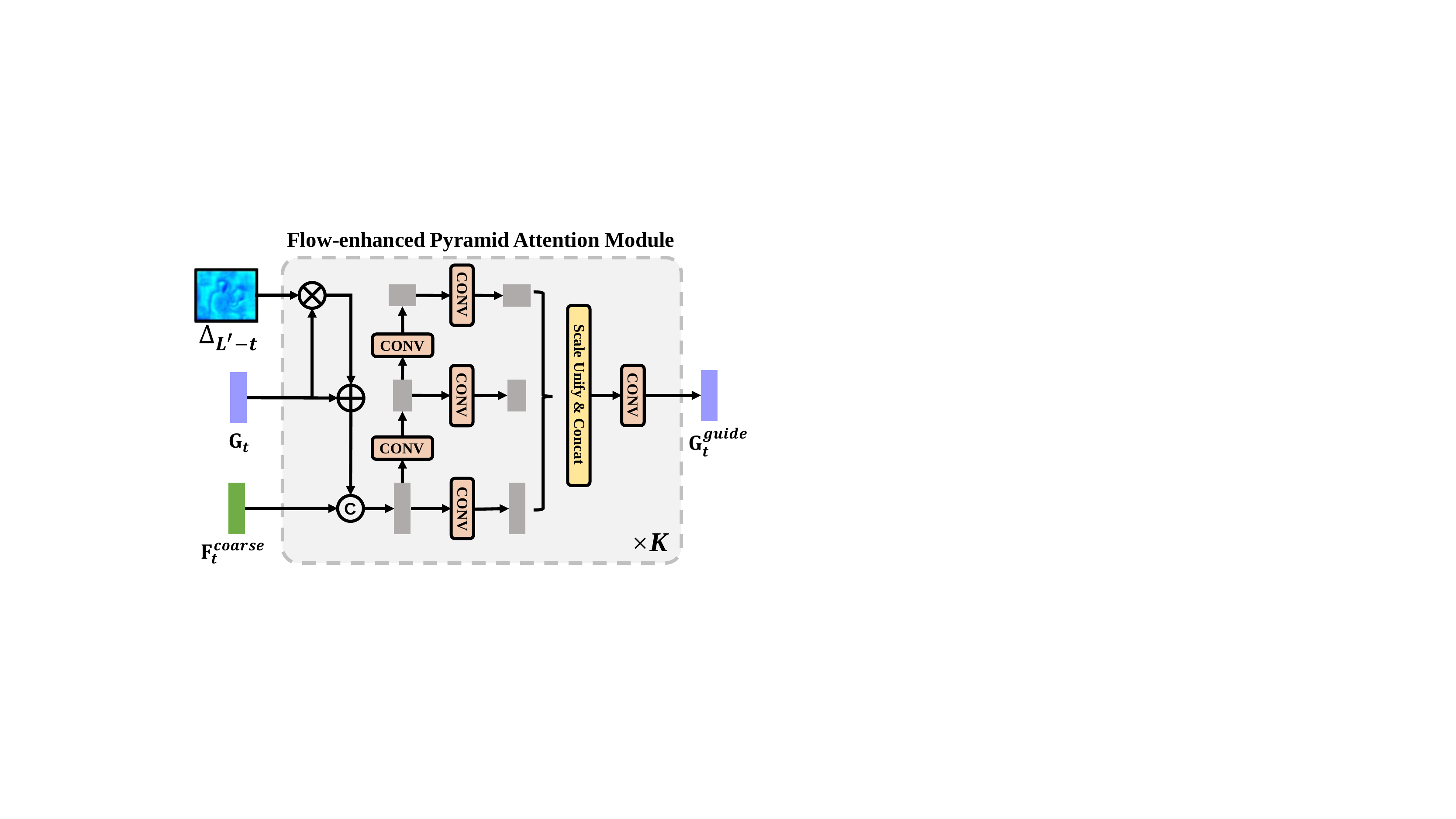}
   \caption{The architecture of the pyramid attention module. The subscript $t$ denotes the feature output in the $t$-th layer of the encoder ($1 \leq t \leq T$). `$\times K$' indicates the iteration times of the guidance feature updating.} 
   \label{fig:Refine}
\end{figure}

\textbf{Flow-enhanced pyramid attention module.} 
In this module, we target at combining the RGB feature and the flow map to learn the edge-focused guidance feature $\{\mathbf{G}_t^{guide}\}$ at each layer. 
In detail, for the $t$-th layer, with the RGB feature $\mathbf{G}_t$ and its corresponding flow map $\mathbf{\Delta}_{L'-t}$, we first fuse the flow information into the RGB feature to form the flow-enhanced RGB feature, 
\begin{equation}\small
\begin{split}
{\mathbf{G}}_{t}^{flow}= \boldsymbol{\Delta}_{L'-t}\cdot{\mathbf{G}}_{t} + {\mathbf{G}}_t,
\end{split}
\end{equation}
where $\boldsymbol{\Delta}_{L'-t}\cdot{\mathbf{G}_t}$ is expected to exploit the significant flow-value fluctuations at the edge region in $\boldsymbol{\Delta}_{L'-t}$ to better highlight the structure region of the RGB feature. 
To further smooth the texture feature in $\mathbf{G}_t^{flow}$, we concatenate it with the texture-less depth feature $\mathbf{F}_t^{coarse}$ to obtain the texture-degraded RGB feature $\Tilde{\mathbf{G}}_t^{flow}$. 
Then, we feed  $\Tilde{\mathbf{G}}_t^{flow}$ into a pyramid network to extract its edge-focused guidance features $\{\Tilde{\mathbf{G}}^{flow}_{t,k}\}_{k=1}^{K}$ at different scales. 
The low-level guidance feature is to filter the texture noise (guided by the flow map) while the high-level is to exploit the rich context information for edge-feature capture. 
After that, we unify the scales of the hierarchical feature $\{\Tilde{\mathbf{G}}_{t,k}^{flow}\}_{k=1}^K$ using the bicubic interpolation and pass the concatenated feature into a convolutional block to generate the flow-enhanced RGB guidance feature $\mathbf{G}_t^{guide}$ at the $t$-th layer. 
Notably, we design an iterative architecture to progressively refine the RGB guidance feature as illustrated in Fig.~\ref{fig:Refine}. 

\textbf{Edge decoder.} 
Guided by the flow-based guidance features $\{\mathbf{G}_t^{guide}\}_{t=1}^T$ learned at each layer, we progressively decode the depth feature in an iterative manner:
\begin{equation}\small
\begin{split}
{\mathbf{F}}_{t+1}^{edge}= \operatorname{FU}({\operatorname{Cat}(\mathbf{F}}_{t}^{edge}, \mathbf{G}_{T-t+1}^{guide}, \mathbf{F}_{T-t+1}^{coarse})),
\end{split}
\end{equation}
where $\operatorname{FU}$ function indicates the fusion and upsampling operation following \cite{guo2020closed} and the initial feature $\mathbf{F}_{1}^{edge}$ is obtained by the convolutional operation on $\mathbf{F}_{T+1}^{coarse}$. 
Finally, we pass $\mathbf{F}_{T+1}^{edge}$ into a convolutional block to obtain the edge-refined HR depth map $\mathbf{D}^{refine}$.

\subsection{Loss Function}
We train our model by minimizing the smooth-L1 loss between 
the ground-truth depth map $\mathbf{D}^{gt}$ and the network output of each sub-network, including the coarse depth prediction $\mathbf{D}^{coarse}$ and the refined one $\mathbf{D}^{refine}$:
\begin{equation}\small
\begin{split}
\mathcal{L}_{dsr} = \sum_{i=1}^{H\times W} \ell\left(\mathbf{D}_i^{coarse} - \mathbf{D}_i^{gt}\right) + \ell\left(\mathbf{D}_i^{refine} - \mathbf{D}_i^{gt}\right),
\end{split}
\end{equation}
where the subscript $i$ denote the pixel index and the smooth-L1 loss function is defined as:
\begin{equation}
\begin{split}
\ell(u)= \begin{cases}0.5 u^{2}, & \text { if }|u| \leq 1 \\ \left(|u|- 0.5\right), & \text { otherwise.}\end{cases}
\end{split}
\end{equation}


\section{Experiments}
\subsection{Experimental Setting}
To evaluate the performance of our method, we perform extensive experiments on
real-world RGB-D-D dataset \cite{he2021towards}, ToFMark dataset \cite{ferstl2013image} and synthetic NYU-v2 dataset \cite{silberman2012indoor}.
We implement our model with PyTorch and conduct all experiments on a server containing an Intel i5 2.2 GHz CPU and a TITAN RTX GPU with almost 24 GB. 
During training, we randomly crop patches of resolution $256\times256$ as groundtruth and the training and testing data are normalized to the range $[0,1]$.
In order to balance the training time and network performance, the parameters $L$, $L'$, $K$, $T$ are set to $3$, $6$, $3$, $2$ in this paper. We quantitatively and visually compare our method with 13 state-of-the-art (SOTA) methods: TGV \cite{ferstl2013image},
FBS \cite{barron2016fast},
MSG \cite{2016Multi},
 DJF \cite{li2016deep},
 DJFR \cite{li2019joint}, GbFT  \cite{albahar2019guided}, PAC \cite{su2019pixel}, CUNet \cite{deng2020deep}, FDKN \cite{kim2021deformable}, DKN \cite{kim2021deformable}, FDSR \cite{he2021towards}, CTKT \cite{sun2021learning} and DCTNet \cite{zhao2022discrete}. 
 For simplicity, we name our \textbf{S}tructure \textbf{F}low-\textbf{G}uided method as \textbf{SFG}.


\begin{figure*}[t]
  \centering
     \includegraphics[width=0.95\linewidth]{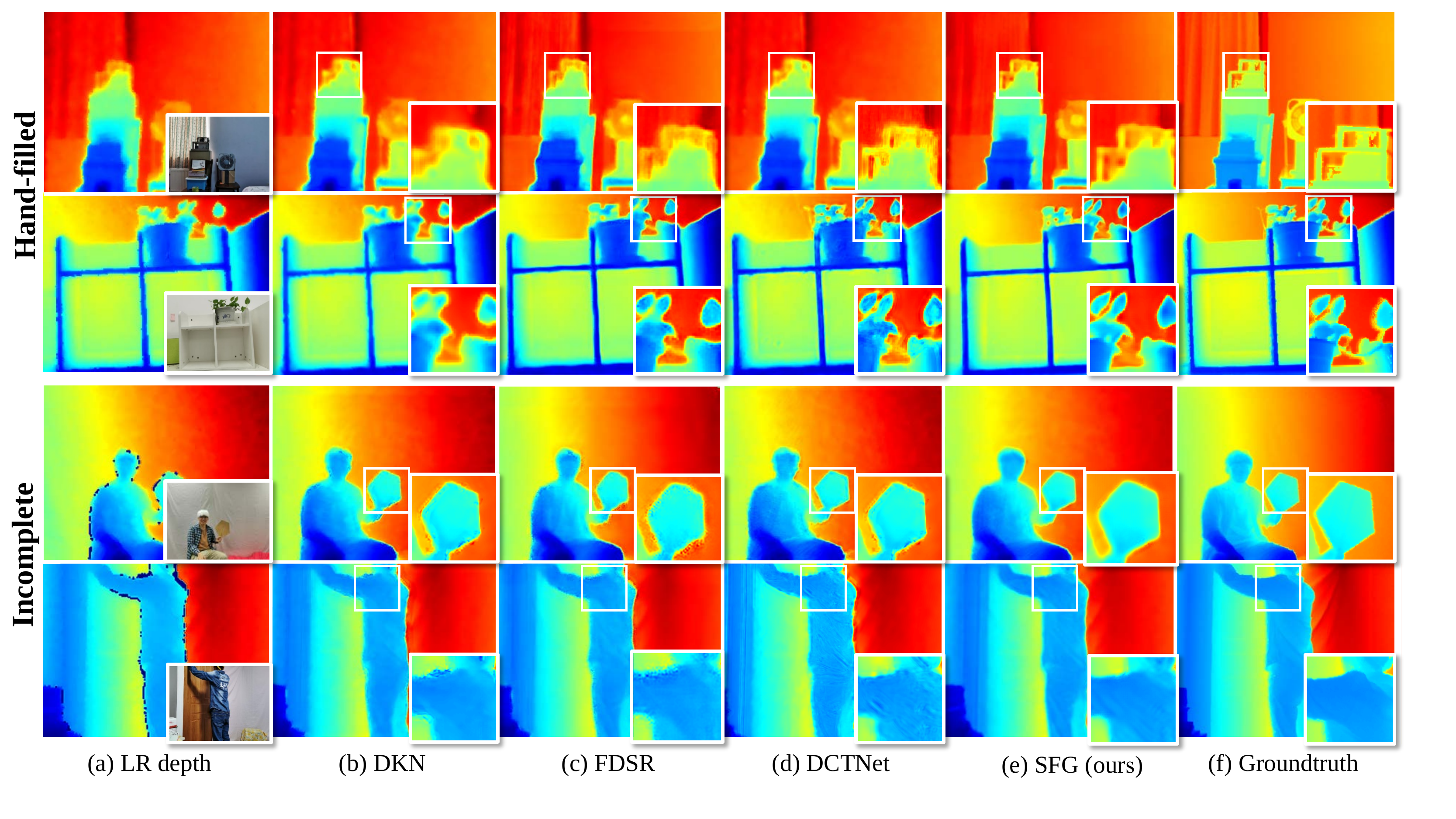}
   \caption{Visual comparison on RGB-D-D dataset. The first (last) two rows show DSR results of hand-filled (incomplete) LR.
   }
   \label{fig:RGBDD}
\end{figure*}
\begin{figure}[t]
  \centering
     \includegraphics[width=0.93\linewidth]{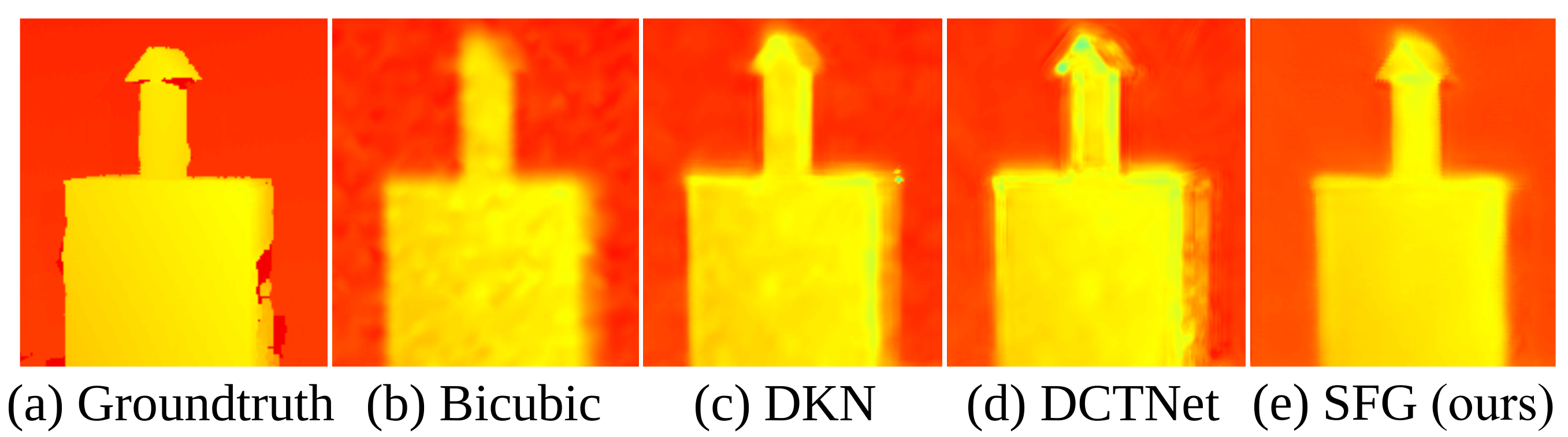}
   \caption{Visual comparison on ToFMark dataset. 
   }
   \label{fig:tof}
\end{figure}
\begin{figure*}[t]
  \centering
     \includegraphics[width=0.85\linewidth]{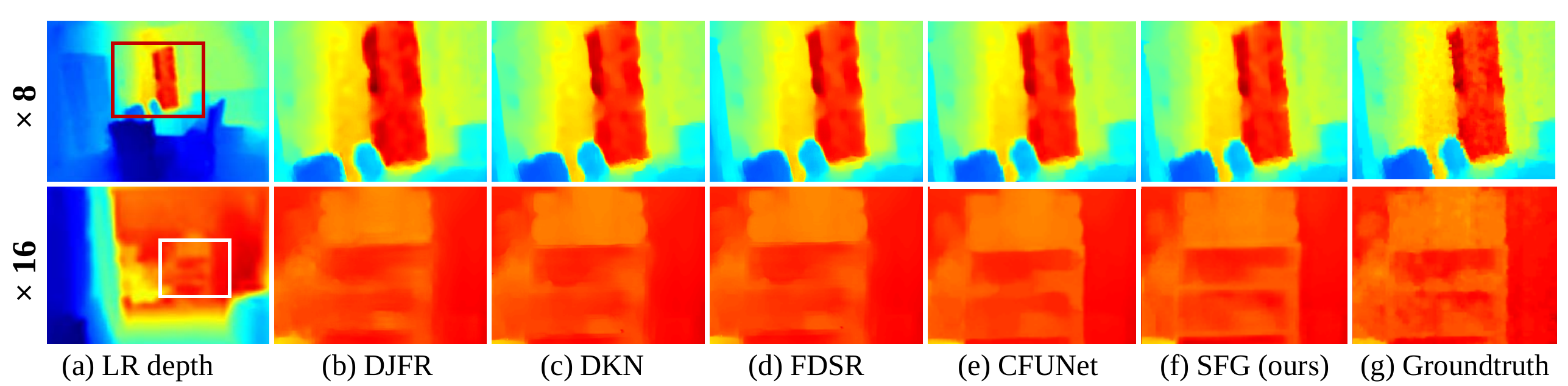}
   \caption{Visual comparison of $\times$8 and $\times$16 DSR results on NYU-v2 dataset.
   }
   \label{fig:NYU}
\end{figure*}

\begin{table*}[t]
\centering
\resizebox{0.87\linewidth}{!}{
\begin{tabular}{l|ccccccccc|c}
\toprule[1.2pt]
RMSE & Bicubic & MSG &DJF& DJFR & CUNet &DKN & FDKN&  FDSR & DCTNet & SFG (ours)\\
\midrule 
Hand-filled&7.17&5.50&5.54&5.52&5.84&\underline{5.08}&5.37&5.34& 5.28 & \textbf{3.88}\\
Incomplete&-&7.90&5.70&5.52&6.54&\underline{5.43}&5.87&5.59& 5.49 &\textbf{4.79}\\
Noisy&11.57&10.36&5.62&5.71&6.13&\underline{5.16}&5.54&5.63& \underline{5.16} & \textbf{4.45}\\
\bottomrule[1.2pt]
\end{tabular}}
\caption{Quantitative comparison on RGB-D-D dataset. Best and second best results are in \textbf{bold} and \underline{underline}, respectively.}\label{tab:rgbd1}
\end{table*}


\begin{table}[t]
	\centering
	\resizebox{1.\linewidth}{!}{
\begin{tabular}{l|ccccc|c}
\toprule[1.2pt]
& DJFR &DKN & FDKN& FDSR & DCTNet & SFG (ours)\\
\midrule
RMSE& 0.27 & \underline{0.26}& 0.28 & 0.28 & 0.27 & \textbf{0.25} \\
\bottomrule[1.2pt]
\end{tabular}}
\caption{Quantitative comparison on ToFMark dataset.}
\label{tab:rgbd2}
\end{table}


\begin{table*}[t]
	\centering
	\resizebox{0.9\linewidth}{!}{
\begin{tabular}{c|ccccccccccc|c}
\toprule[1.2pt]
RMSE&TGV & FBS & DJFR & GbFT & PAC &CUNet& FDKN  & DKN &FDSR  &DCTNet  & CTKT&  SFG (ours)\\
\midrule
$\times4$ &4.98&4.29  &2.38  &3.35  & 2.39 &1.89& 1.86 & 1.62 & 1.61  & 1.59&\underline{1.49} &\textbf{1.45}\\
$\times8$&11.23&8.94  &4.94  &5.73  & 4.59 & 3.58&3.33  & 3.26 & 3.18  & 3.16 &\textbf{2.73}& \underline{2.84}\\
$\times16$ &28.13&14.59  &9.18 &9.01 & 8.09 & 6.96&6.78 & 6.51 & 5.86  & 5.84 & \textbf{5.11} & \underline{5.56} \\
\bottomrule[1.2pt]
\end{tabular}}
\caption{Quantitative comparison on NYU-v2 dataset in terms of average RMSE (cm).}
\label{tab:NYU-v2}
\end{table*}



\subsection{Experiments on Real Datasets}
Depth maps captured by cheap depth sensors usually suffer from structural distortion and edge noise. To verify the efficiency and robustness of our proposed method, we employ our method on two challenging benchmarks: RGB-D-D dataset and ToFMark dataset.

\noindent\textbf{Evaluation on hand-filled RGB-D-D.} 
To evaluate the performance of our method on real LR depth maps, we conduct experiments on RGB-D-D datasets captured by two RGB-D sensors: Huawei P30 Pro (captures RGB images and LR depth maps) and Helios TOF camera (captures HR depth maps).
The LR inputs are shown in Fig.~\ref{fig:RGBDD}, which suffer from the low resolution (LR size is $192 \times 144$ and target size is $512 \times 384$) and random structural missing in the edge region. 
Following FDSR \cite{he2021towards}, we first use 2215 hand-filled RGB/D pairs for training and 405 RGB/D pairs for testing.
As listed in the first row of Table~\ref{tab:rgbd1}, the proposed model outperforms SOTA methods by a significant margin. 

The first two rows in Fig.~\ref{fig:RGBDD} show the visual DSR comparisons on hand-filled RGB-D-D dataset. We can see that edges in the results of DKN \cite{kim2021deformable} and DCTNet \cite{zhao2022discrete} are over-smoothed and the artifacts are visible in the FDSR results.
In contrast, our results show more accurate structures without texture copying.

\noindent\textbf{Evaluation on incomplete RGB-D-D.} 
To further verify the DSR performance of our method in the case of edge noise (e.g., edge holes), instead of the hole completion above, we directly test SFG on unfilled RGB-D dataset and achieve the lowest RMSE as shown in the second row of Table~\ref{tab:rgbd1}. 
Moreover, as shown in the last two rows in Fig.~\ref{fig:RGBDD}, the edges recovered by our method are sharper with fewer artifacts and visually closest to the ground-truth map. 
It's mainly attributed to the edge-focused guidance feature learning with our flow-enhanced pyramid edge attention network.

\noindent\textbf{Evaluation on noisy RGB-D-D and ToFMark.} 
We evaluate the denoising and generalization ability of our method on ToFMark dataset consisting of three RGB-D pairs. The LR inputs have irregular noise and limited resolution (LR depth is $120 \times 160$ and target size is $610 \times 810$). 
To simulate the similar degradation for training, we add the Gaussian noise (mean 0 and standard deviation 0.07) and the Gaussian blur (kernel size 5) on the 2215 RGB-D pairs from RGB-D-D dataset to generate the noisy training dataset. Testing dataset consists of 405 RGB-D pairs from noisy RGB-D-D dataset and 3 RGB-D pairs from ToFMark dataset.
As shown in the last row of Table~\ref{tab:rgbd1} and Table~\ref{tab:rgbd2}, our method achieves the lowest RMSE in noisy RGB-D-D dataset and the lowest RMSE in ToFMark dataset, which proves its ability for noise removing. 
As shown in Fig.~\ref{fig:tof}, it is observed that DKN \cite{kim2021deformable} and DCTNet \cite{zhao2022discrete} introduce some texture artifacts and noise in the low-frequency region, while SFG recovers clean surface owing to PEA with effective texture removing.

\subsection{Experiments on Synthetic Datasets}
Since most popular methods are designed for synthetic datasets, we further evaluate our method on NYU-v2 datasets for a more comprehensive comparison. Following the widely used data splitting criterion, we sample 1000 RGB-D pairs for training and the rest 449 RGB-D pairs for testing. 
As shown in the Table~\ref{tab:NYU-v2}, the proposed method still achieves comparable results with the SOTA methods on all upsampling cases ($\times4, \times8, \times16$). 
In addition, Fig.~\ref{fig:NYU} presents that our $\times8$ and $\times16$ upsampled depth maps own higher accuracy and more convincing results. It verifies that our method not only performs DSR well in low-quality maps with noise and missing structure, but also achieves high-quality precision in the case of large-scale upsampling. 
\begin{table}[t]
	\centering
	\resizebox{0.62\linewidth}{!}{
\begin{tabular}{l|c}
\toprule[1.2pt]
Model&\multicolumn{1}{|c}{RMSE}\\
\midrule
CFUNet &
\multicolumn{1}{|c}{4.22}\\
CFUNet \textit{w/o} TriSA&
\multicolumn{1}{|c}{4.34}\\
CFUNet \textit{w/o} cross-attention&
\multicolumn{1}{|c}{4.57}\\
\bottomrule[1.2pt]
\end{tabular}}
\caption{Ablation study of CFUNet on RGB-D-D dataset.}
\label{tab:Ablation1}
\end{table}

\begin{table}[t]
	\centering
	\resizebox{0.70\linewidth}{!}{
\begin{tabular}{l|cc}
\toprule[1.2pt]
Datasets & SFG & SFG \textit{w/o} PEANet\\
\midrule
RGB-D-D & 3.88 & 4.22 \\
NYU-v2 ($\times4$) &1.45& 1.82 \\
NYU-v2 ($\times8$)  &2.84&3.76\\
NYU-v2 ($\times16$) &5.55& 5.90\\
\bottomrule[1.2pt]
\end{tabular}}
\caption{Ablation study (in RMSE) of PEANet.}
\label{tab:Ablation2}
\end{table}
\subsection{Ablation Analysis}
\noindent\textbf{Ablation study on CFUNet.}
As shown in the first row of the Table~\ref{tab:Ablation1}, we still achieve the lowest RMSE criterion just with the single CFUNet (SFG w/o PEANet) on RGB-D-D dataset when compare with SOTA methods. It proves the effectiveness of the learned structure flow map for real DSR. 
The Table~\ref{tab:Ablation1} also shows that removing the trilateral self-attention (TriSA) and cross-attention module in CFUNet causes performance degradation on RGB-D-D datasets, which verifies the necessary of the depth feature enhancement for reliable flow map generation.

\begin{figure}[t]
  \centering
     \includegraphics[width=0.90\linewidth]{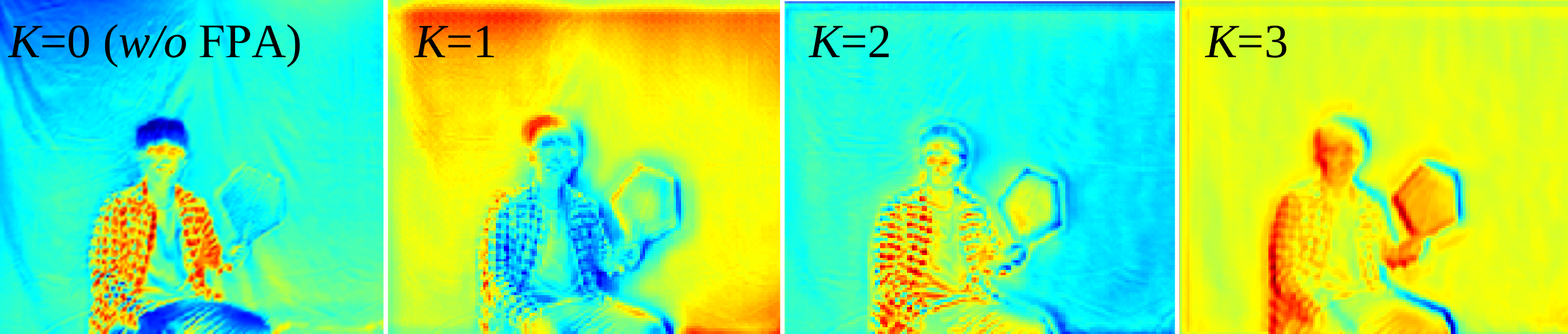}
   \caption{Visual comparison of guidance features using FPA with different iteration times $\textit{K}$, \emph{i.e.}, from 0 (\textit{w/o} FPA) to 3.}
   \label{fig:Ab_it}
\end{figure}

\noindent\textbf{Ablation study on PEANet.}
To analyze the effectiveness of PEANet, we train the network with and without PEANet on the synthetic dataset (NYU-v2) and the real-world dataset (RGB-D-D). As shown in the Table~\ref{tab:Ablation2}, PEANet consistently brings the RMSE gain under both real and synthetic dataset settings. 
It's mainly due to our edge-focused guidance feature learning for robust edge refinement. 
In addition, Fig.~\ref{fig:Ab_it} shows the guidance features under varying iteration times in FPA (\textbf{F}low-enhanced \textbf{P}yramid \textbf{A}ttention) module from 0 (\textit{w/o} FPA) to 3. 
Visually, as the number of iterations increases, the edge regions tend to receive more attention.

\section{Conclusion}
In this paper, we proposed a novel structure flow-guided DSR framework for real-world depth super-resolution, which deals with issues of structural distortion and edge noise. For the structural distortion, a cross-modality flow-guided upsampling network was presented to learn a reliable cross-modality flow between depth and the corresponding RGB guidance for the reconstruction of the distorted depth edge, where a trilateral self-attention combines the geometric and semantic correlations for structure flow learning. 
For the edge noise, a flow-enhanced pyramid edge attention network was introduced to produce edge attention based on the learned flow map and learn the edge-focused guidance feature for depth edge refinement with a pyramid network. Extensive experiments on both real-world and synthetic datasets demonstrated the superiority of our method.

\section{Acknowledgement}
This work was supported by the National Science Fund of China under Grant Nos.~U1713208 and 62072242.

\bibliography{aaai23}

\end{document}